\documentclass[runningheads]{llncs}
\usepackage{amsmath,graphicx}
\usepackage{booktabs}
\usepackage{multirow}
\usepackage{caption}
\usepackage{subcaption}

\begin{document}

\title{Evaluating Sugarcane Yield Variability with UAV-Derived Cane Height under Different Water and Nitrogen Conditions}

\author{
Rajiv Ranjan \orcidID{0009-0008-3248-7374} \and 
Tejasavi Birdh \and 
Nandan Mandal\ \and 
Dinesh Kumar\ \and 
Shashank Tamaskar\
}

\authorrunning{R. Ranjan et al.}

\institute{
Plaksha University, Punjab, India\\
\email{\{rajiv.ranjan, tejasavi.birdh, nandan.mandal, dinesh.kumar1, shashank.tamaskar\}@plaksha.edu.in}
}

\maketitle

\begin{abstract}
This study investigates the relationship between sugarcane yield and cane height derived under different water and nitrogen conditions from pre-harvest Digital Surface Model (DSM) obtained via Unmanned Aerial Vehicle (UAV) flights over a sugarcane test farm. The farm was divided into 62 blocks based on three water levels (low, medium, and high) and three nitrogen levels (low, medium, and high), with repeated treatments. In pixel distribution of DSM for each block, it provided bimodal distribution representing two peaks, ground level (gaps within canopies) and top of the canopies respectively.  Using bimodal distribution,  mean cane height was extracted for each block by applying a trimmed mean to the pixel distribution, focusing on the top canopy points. Similarly, the extracted mean elevation of the base was derived from the bottom points, representing ground level. The Derived Cane Height Model (DCHM) was generated by taking the difference between the mean canopy height and mean base elevation for each block. Yield measurements (tons/acre) were recorded post-harvest for each block. By aggregating the data into nine treatment zones (e.g., high water-low nitrogen, low water-high nitrogen), the DCHM and median yield were calculated for each zone. The regression analysis between the DCHM and corresponding yields for the different treatment zones yielded an $R^{2}$ of 0.95. This study demonstrates the significant impact of water and nitrogen treatments on sugarcane height and yield, utilizing one-time UAV-derived DSM data.
\end{abstract}

\section{Introduction}
\label{sec:intro}
Precision agriculture has revolutionized farming practices by integrating advanced technologies to enhance crop management and productivity. Among these technologies, the use of Unmanned Aerial Vehicles (UAVs) has gained significant attention due to their ability to capture high-resolution aerial imagery, providing valuable data for crop monitoring and yield estimation. Sugarcane, a critical crop for many economies, serves as a primary source of sugar and bio-fuel. Accurate yield estimation is essential for effective resource management and optimizing production. Traditional methods of yield estimation such as crop cutting experiments are often labor-intensive and time-consuming. UAV technology offers a more efficient and precise alternative, enabling farmers to monitor crop growth and predict yields with greater accuracy.

Digital Surface Models (DSMs), representing the Ea  surface including vegetation, are crucial in this context. They are used to measure crop heights, a critical parameter for assessing crop health and estimating yield \cite{rs9030239}. One significant area of research is the use of UAVs for creating DSMs to generate canopy height models (CHM) \cite{CHANG2017232} and monitor crop growth. For instance, Li et al. \cite{LI2016637} and Harkel et al. \cite{ten2019biomass} demonstrated the application of UAV-derived heights in estimating the Above Ground Biomass (AGB) of maize and other crops, showing a good correlation between UAV-derived heights and actual biomass measurements. Similarly, Wang et al. utilized UAVs to generate DSMs for wheat \cite{wang2022estimation}, achieving accurate crop height measurements that correlated well with ground truth data. Similarly, many analyses have been done using UAV-derived height of crops like sorghum \cite{10.3389/fpls.2017.00421}, corn \cite{furukawa2020corn}, barley \cite{rs61110395}, maize \cite{agriculture10050146}, wheat \cite{ecrs-2-05163}, cotton \cite{rs15051214}, and sugarcane \cite{de2017height}.

In the context of sugarcane, studies have shown promising results using remote sensing technologies for sugarcane canopy detection. Various studies have highlighted the advantages of satellite and UAV-based remote sensing over traditional methods, emphasizing its efficiency, accuracy, and cost-effectiveness. Shendryk et al. \cite{shendryk2021integrating} employed satellite imagery to estimate field level sugarcane yield, while vargas et al. explored the use of LiDAR data for similar purposes \cite{vargas2024estimating}. Ranjan et al. \cite{10640589} presented the study to predict the field level sucrose content (pol value) in sugarcane and optimal harvest date using time series satellite data.  Maes et al. presented the effectiveness of UAVs in monitoring growth stages, assessing plant health, and estimating biomass and yield \cite{MAES2019152}. Chea et al. used high-resolution UAV images and DSMs to detect the sugarcane canopy in a field \cite{chea2019sugarcane}. In sugarcane farming, UAVs have proven to be highly beneficial. They are used to monitor growth, assess plant health, and estimate biomass and yield \cite{chiranjeeb2022application}\cite{shikhar2024evaluation}.  These studies underscored the potential of remote sensing in providing timely and accurate yield estimates, although UAV-based approaches were not extensively covered. Building upon these foundational studies, our research employs UAV-derived DSMs to directly estimate sugarcane yield. In this work, the integration of UAVs in precision agriculture has been extensively researched, primarily due to their capability to capture high-resolution imagery for crop monitoring and yield estimation. By capturing DSM images at key growth stages and analyzing height variations is very efficient for the study, this study offers a novel approach that enhances the precision and reliability of yield predictions in sugarcane farming. We employed a Phantom 4 UAV to capture DSM images of a sugarcane farm just before harvesting. This DSM represented the crop at its optimal height. By analysing the DSM data with statistical method like pixel distribution, we obtained a detailed unimodel or bimodal distribution representing denser canopy cover and sparser canopy cover respectively. The primary objectives were to assess the impact of water and nitrogen levels on the cane's height and eventually on its yield. The reason for assessing water and nitrogen stress in sugarcane is that, these factors are the primary contributors to yield gaps in sugarcane production before crop management practices \cite{DIAS20181127}. A yield gap is the difference between the potential yield of a crop, achieved under optimal conditions with maximum resource use and management, and the actual yield obtained by farmers under real-world conditions. The study area was divided into 62 small blocks, each subjected to different levels of water and nitrogen (fertilizer) treatment, and recorded yield values in tons per acre. We computed the trimmed mean cane height for each block by applying block-specific masks to the DSM image. A regression analysis was then performed to investigate the relationship between surrogate sugarcane derived height and yield, revealing a strong correlation with an R² value of 0.95. This paper aims to demonstrate the effectiveness of UAV-derived DSMs in estimating sugarcane yield and to highlight the potential of this technology for enhancing precision agriculture practices. 

\section{Experimental Setup}

\subsection{Study Area and Data Collection}
\label{ssec:subhead}

\begin{figure}[th!]
    \centering
    \includegraphics[width=\linewidth]{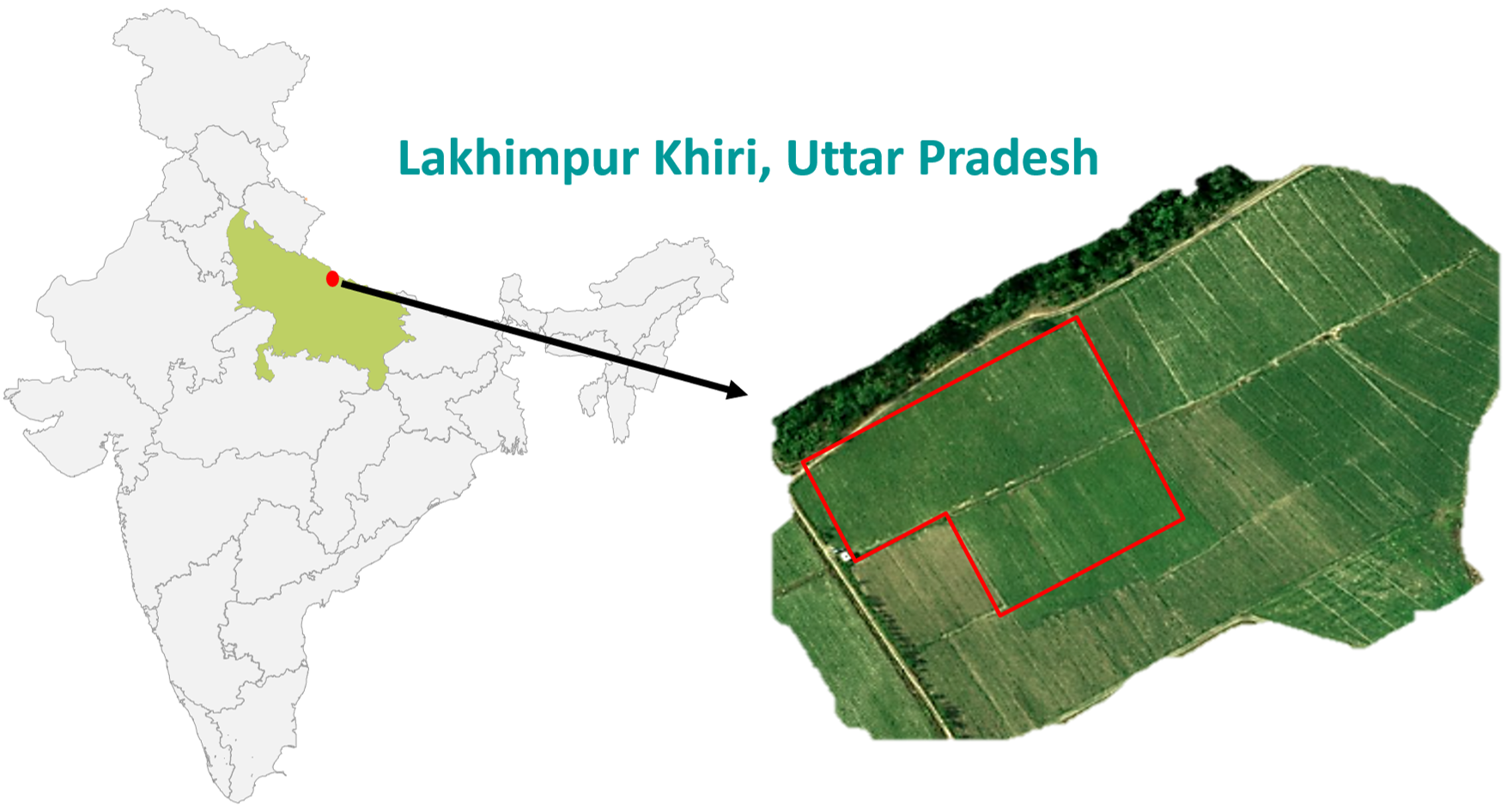}
    \caption{Test farm inside the red boundary at Lakhimpur Khiri, Uttar Pradesh, India}
    \label{fig:roi}
\end{figure}

Uttar Pradesh (India) is the main cane-growing state in the country, allocating about 2.2 million ha (43.7\%) of its area for cane cultivation \cite{sharma2017sugarcane}. We established a 3-acre test farm (Figure~\ref{fig:roi}) in Lakhimpur Kheri District (\(28^\circ 27' 14.2''\text{N} \quad \allowbreak 80^\circ 55' 42.0''\text{E}\)), Uttar Pradesh, India, situated in the upper Gangetic plains agro-climatic zone with rains in a subtropical region. UAV data and ground truth measurements were collected at regular intervals (monthly) throughout the crop cycle, from February 2023 to February 2024. However, for this study, only the digital surface model (DSM) from 24th February 2024 was considered, just before the crop was harvested on 25th February 2024.

\subsection{Farm Layout}
\label{sssec:farm layout}
\begin{figure}[t!]
    \centering
    \includegraphics[width= 1 \linewidth]{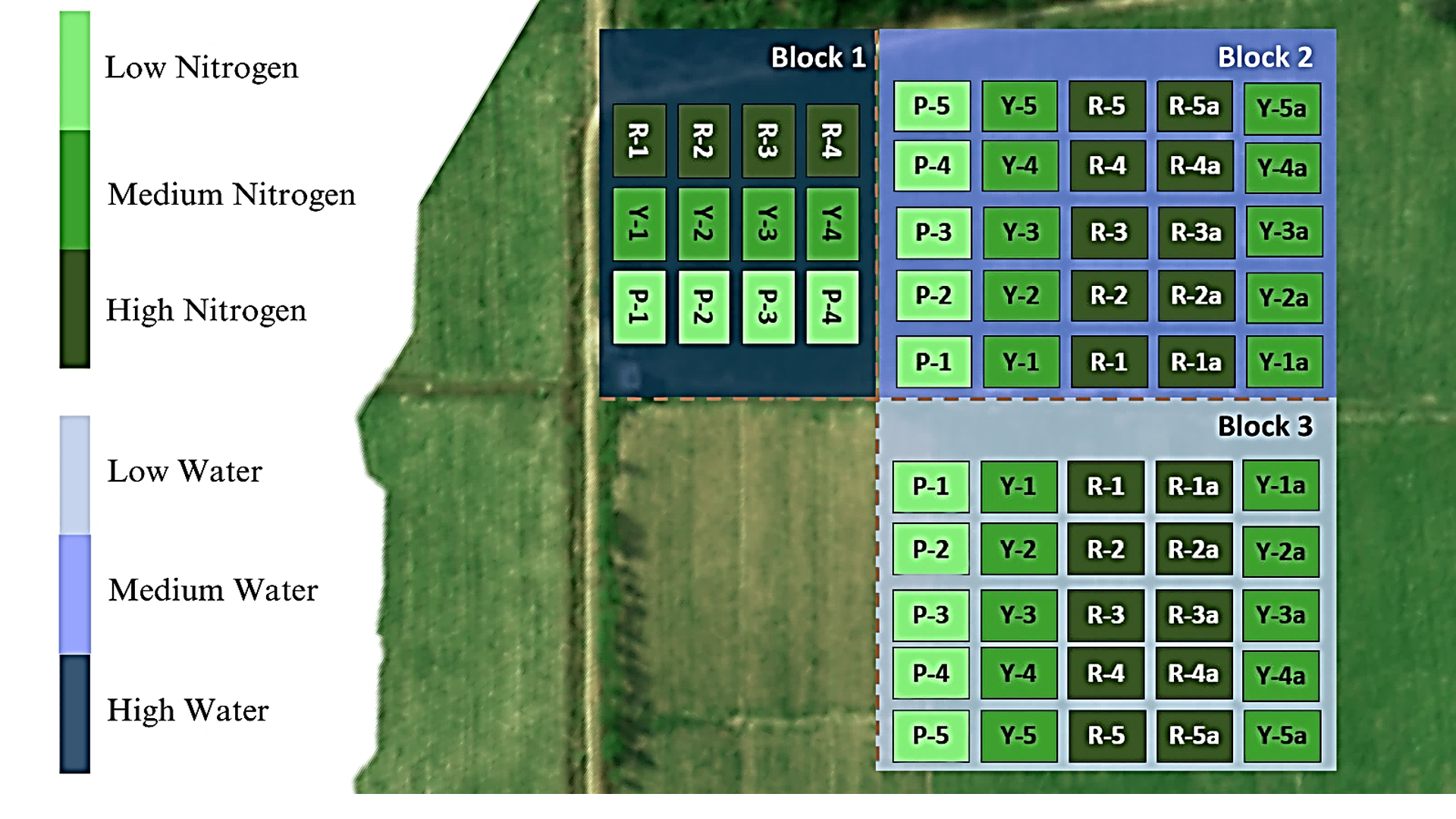}
    \caption{Farm layout}
    \label{fig:layout}
\end{figure}

The test farm is divided into three zones based on water application levels designated as LW (Low Water), MW (Medium Water), and HW (High Water) represented by different shades of blue. Within each water zone, there are blocks designated for low, medium, and high nitrogen levels represented as LN, MN, and HN respectively, indicated by different shades of green. For nitrogen stress estimation, LN, MN, and HN blocks received 0.25kg, 0.5kg, and 1kg of fertilizer respectively for HW blocks, and 0.2kg, 0.4kg, and 0.8kg for LW and MW blocks. The water levels applied were 50\%, 100\%, and 150\% of the typical application for the LW, MW, and HW water blocks respectively. These amounts were determined based on the recommendations of an agronomist. Monthly drone flights were conducted over the test farm to collect aerial images for analysis. Eventually, 62 small blocks curated as shown in figure \ref{fig:layout}. An appropriate naming convention was used to identify each block. However, a combination of water and nitrogen blocks provides 9 treatment zones based on the treatment level of water and nitrogen explained in Table \ref{table:water_nitrogen_levels}.

\begin{table*}[htbp]
\centering
\caption{Water and Nitrogen Levels Applied to Sugarcane Test Farm with Naming Conventions}
\resizebox{\textwidth}{!}{%
\begin{tabular}{|c|c|c|c|c|}
\hline
\textbf{Water Zone} & \textbf{Water Level} & \textbf{Nitrogen Zone} & \textbf{Fertilizer Applied} & \textbf{Abbreviation} \\ \hline

\multirow{3}{*}{LW (Low Water)}   
& \multirow{3}{*}{50\%} 
& LN (Low Nitrogen)   & 0.2 kg  & LW\_LN \\ \cline{3-5}
&                       
& MN (Medium Nitrogen) & 0.4 kg  & LW\_MN \\ \cline{3-5}
&                       
& HN (High Nitrogen)   & 0.8 kg  & LW\_HN \\ \hline

\multirow{3}{*}{MW (Medium Water)} 
& \multirow{3}{*}{100\%} 
& LN (Low Nitrogen)   & 0.2 kg  & MW\_LN \\ \cline{3-5}
&                       
& MN (Medium Nitrogen) & 0.4 kg  & MW\_MN \\ \cline{3-5}
&                       
& HN (High Nitrogen)   & 0.8 kg  & MW\_HN \\ \hline

\multirow{3}{*}{HW (High Water)}   
& \multirow{3}{*}{150\%} 
& LN (Low Nitrogen)   & 0.25 kg & HW\_LN \\ \cline{3-5}
&                       
& MN (Medium Nitrogen) & 0.5 kg  & HW\_MN \\ \cline{3-5}
&                       
& HN (High Nitrogen)   & 1 kg    & HW\_HN \\ \hline
\end{tabular}%
}
\label{table:water_nitrogen_levels}
\end{table*}

\subsection{Drone Flight}
\label{sssec:subsubhead}
\begin{figure}[h!]
    \centering
    \includegraphics[width= 1 \linewidth]{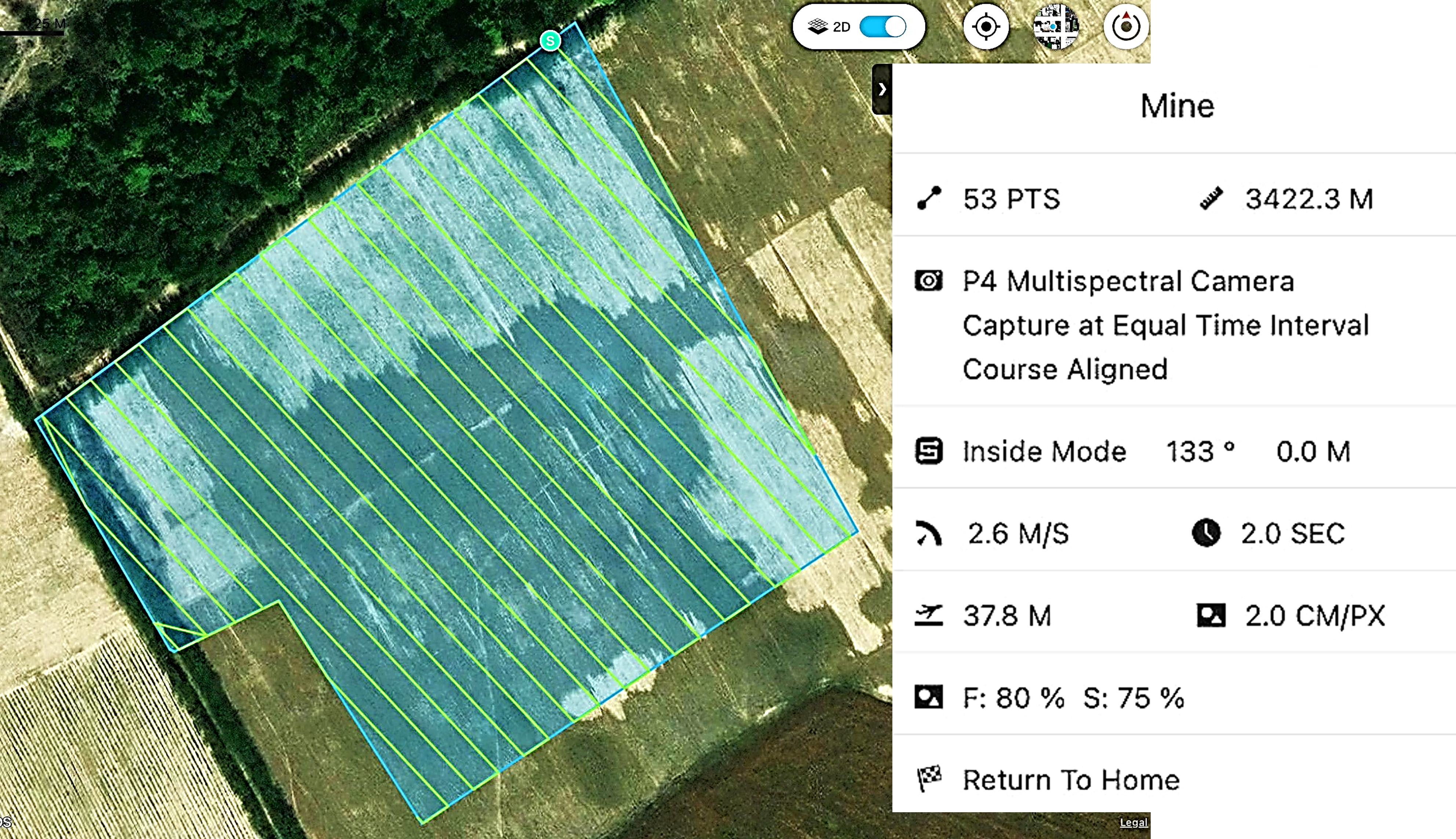}
    \caption{UAV flight mission planning using DJI GS Pro}
    \label{fig:gsp}
\end{figure}

A single DSM image captured on 376 days after planting through a DJI Phantom 4 mounted P4 multispectral camera \cite{Sadenova_Beisekenov_Anuarbekov_Kapasov_Kulenova_2023} was utilized. DJI GS Pro \cite{rs12030351} was used for drone flight mission planning as shown in Figure \ref{fig:gsp}. This UAV was flown at two altitudes of 37.8m and 75.6m generating a spatial resolution (GSD - Ground sample distance) of 2 cm/px. and 4 cm/px. respectively. Front and side overlapping were kept at 80\% and 75\% respectively.

\subsection{Data Preprocessing}
\label{sssec:subsubhead}
\begin{figure}[h!]
    \centering
    \includegraphics[width=1 \linewidth]{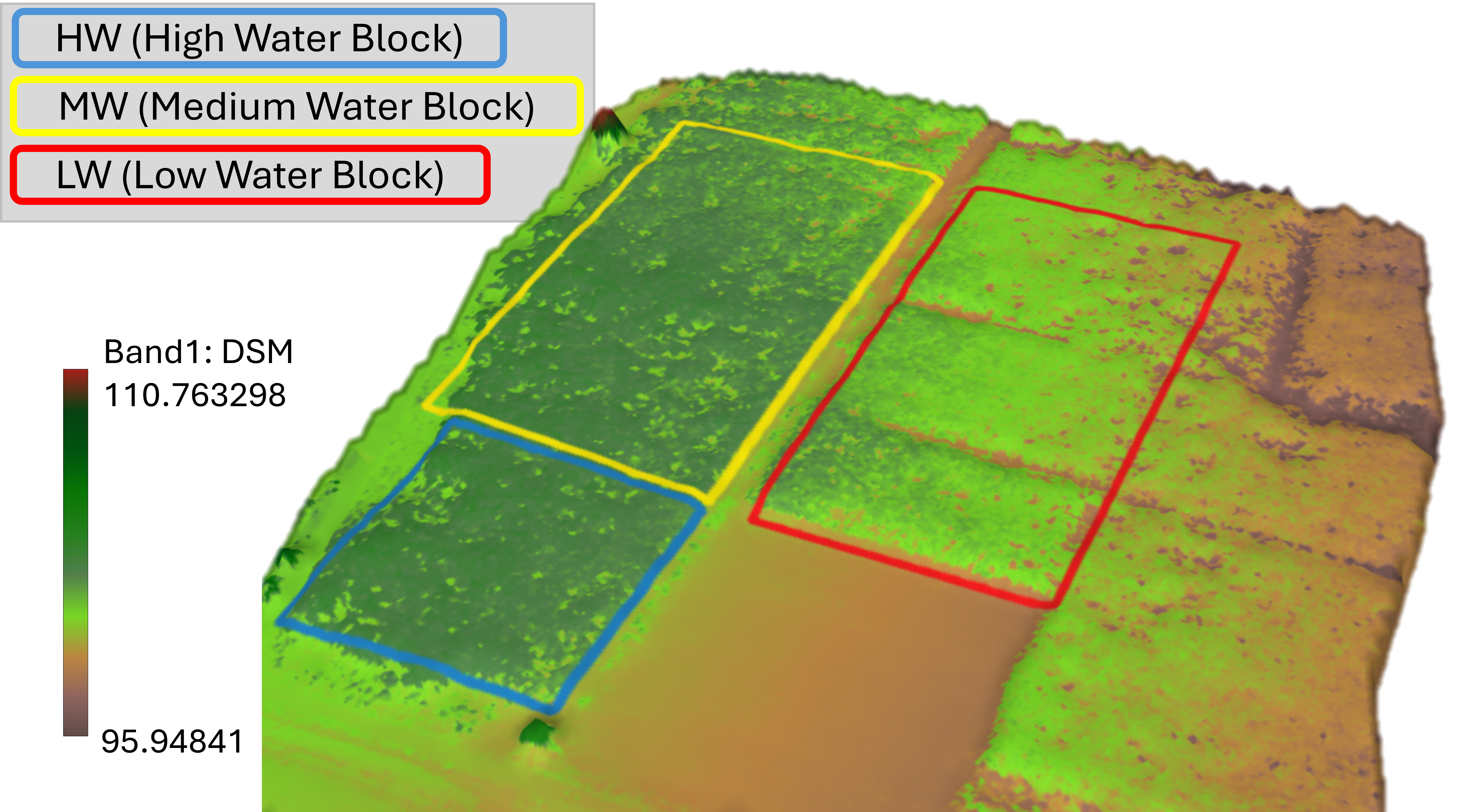}
    \caption{DSM (Digital Surface Model) of the test farm}
    \label{fig:dsm}
\end{figure}

Individual patches captured by a drone are stitched together to create a comprehensive orthomosaic DSM image. This process is essential for accurately mapping and analyzing the farm's terrain and vegetation. We utilized Pix4Dfields software, a powerful tool specifically designed for agricultural applications, to achieve this \cite{10600556}. It aligned and merged the drone-captured images, ensuring high precision and consistency. The resultant orthomosaic DSM image as shown in Figure \ref{fig:dsm} provides an accurate representation of the farm's surface, which is critical for subsequent analysis, including crop height measurement, yield estimation, and stress detection. This high-resolution image enables us to make informed decisions based on detailed visual and topographical data, ultimately enhancing the efficiency and effectiveness of our agricultural practices.

\section{Methodology}
\label{sec:pagestyle}
We utilized a UAV-derived Digital Surface Model (DSM) of a farm, where sugarcane was cultivated. The DSM was captured shortly before the harvest period from multiple altitudes, with the analysis focused on imagery obtained at a $37.8$ meter height. The farm was segmented as discussed in section \ref{sssec:farm layout} for detailed analysis, each of which was subjected to varying water and nitrogen treatments. Upon examining the histograms for each of the 62 plots, a bimodal distribution was observed in most cases. The initial peak on the x-axis indicated the presence of ground pixels within the DSM. This was corroborated by observations that plots situated at the field's periphery, which contained more ground and less crop coverage, exhibited a pronounced initial peak. Conversely, plots located centrally within the crop-planted area showed a diminished first peak, corresponding to gaps in the field as confirmed by ortho-imagery. In rare instances where the first peak was absent, the plots contained very dense sugarcane stands. To quantify the sugarcane height, we considered two distinct scenarios:

\subsection{Case: 1}
\begin{figure}[h!]
    \centering
    \includegraphics[width=0.9\textwidth]{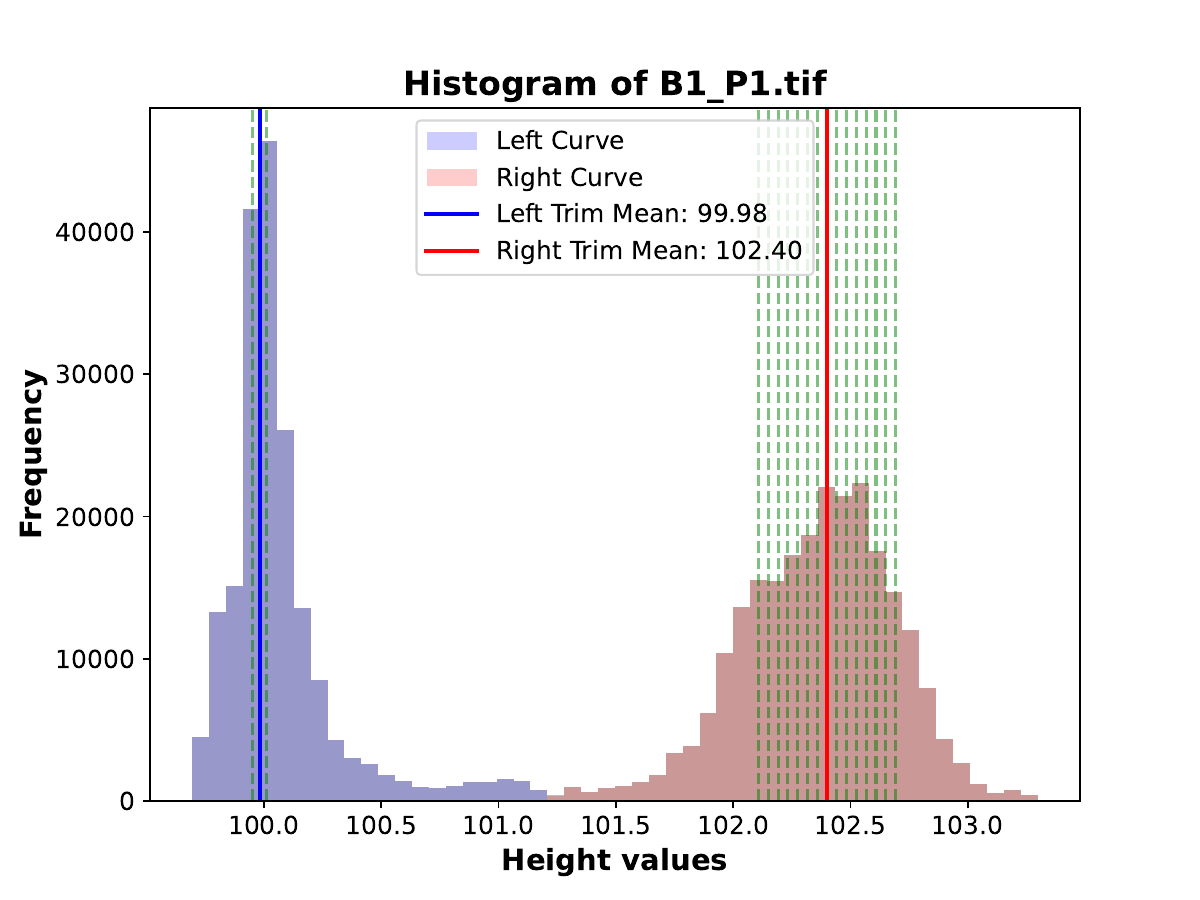}
    \caption{Case 1a - Histogram of Block (B1\_P1) showing a bimodal distribution}
    \label{fig:bimodal1}
\end{figure}

\begin{figure}[h!]
\centering
\includegraphics[width=0.9\textwidth]{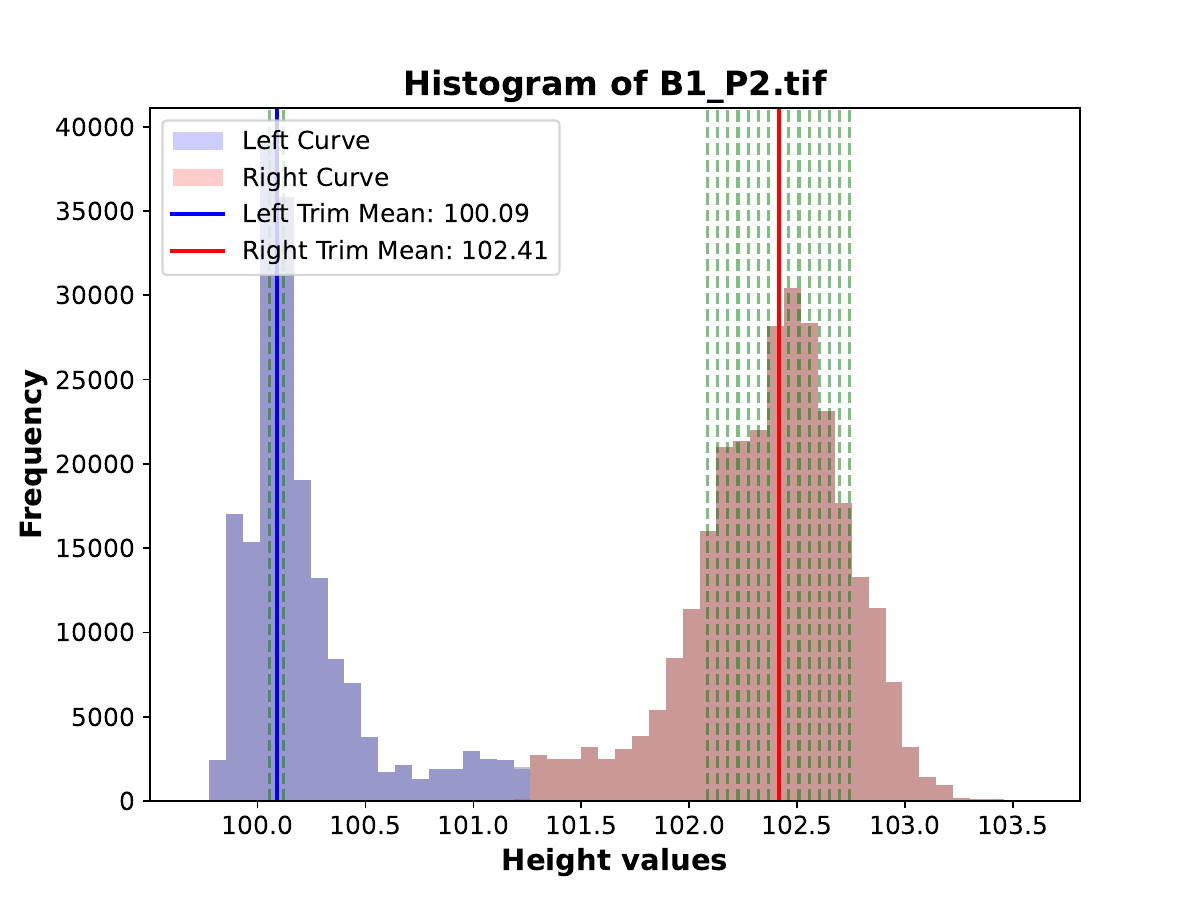}
\caption{Case 1b - Histogram of Block (B1\_P2) showing a bimodal distribution}
\label{fig:bimodal2}
\end{figure}

\begin{figure}[h!]
\centering
\includegraphics[width=1\textwidth]{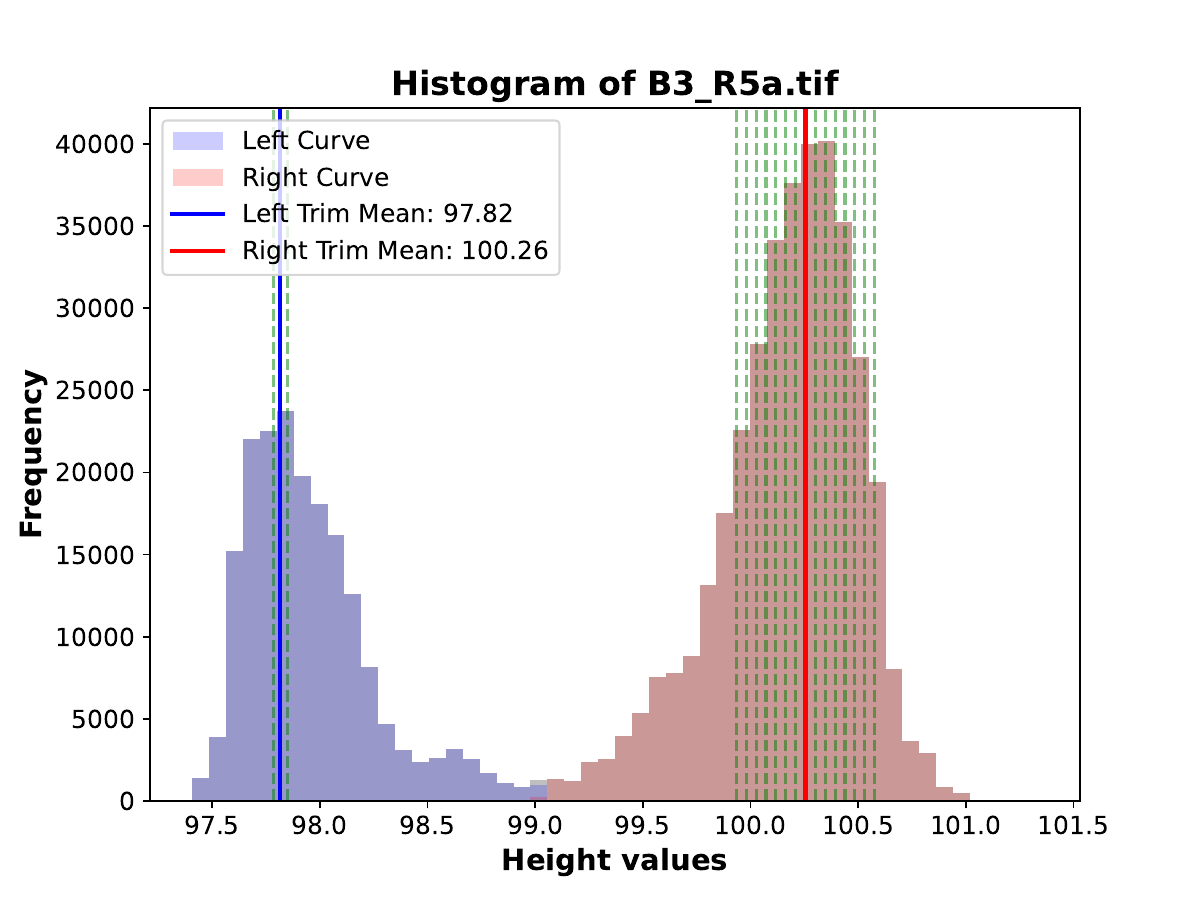}
\caption{Case 1c - Histogram of Block (B3\_R5a) showing a bimodal distribution}
\label{fig:bimodal3}
\end{figure}

To analyze the bimodal distribution observed in the histograms of the majority of the blocks, we employed a Gaussian Mixture Model (GMM) \cite{SKAKUN2017244} to separate the data into two distinct Gaussian curves as shown in figure \ref{fig:bimodal1},\ref{fig:bimodal2} and \ref{fig:bimodal3}. These two curves correspond to different components of the histogram, with one representing the crop height (right curve) and the other representing the ground level (left curve). The GMM fitting allowed us to isolate and model each curve independently. Once the curves were separated, we computed the trimmed mean for each component to obtain more robust estimates that are less sensitive to outliers. For the curve representing the crop height (right curve), we selected the data points corresponding to the highest $30\%$ of frequency values, ensuring that the trimmed mean calculation focused on the most significant portion of the crop height distribution. Similarly, for the curve representing the ground level (left curve), we selected the highest 10\% of frequency values to compute the trimmed mean, capturing the most prominent ground-level data. The difference between these two trimmed means provided a reliable estimate of the surrogate sugarcane height, as it effectively accounted for the height difference between the crop and the ground, minimizing the influence of noise and outliers.

\begin{figure}[h!]
    \centering
    \includegraphics[width=1\textwidth]{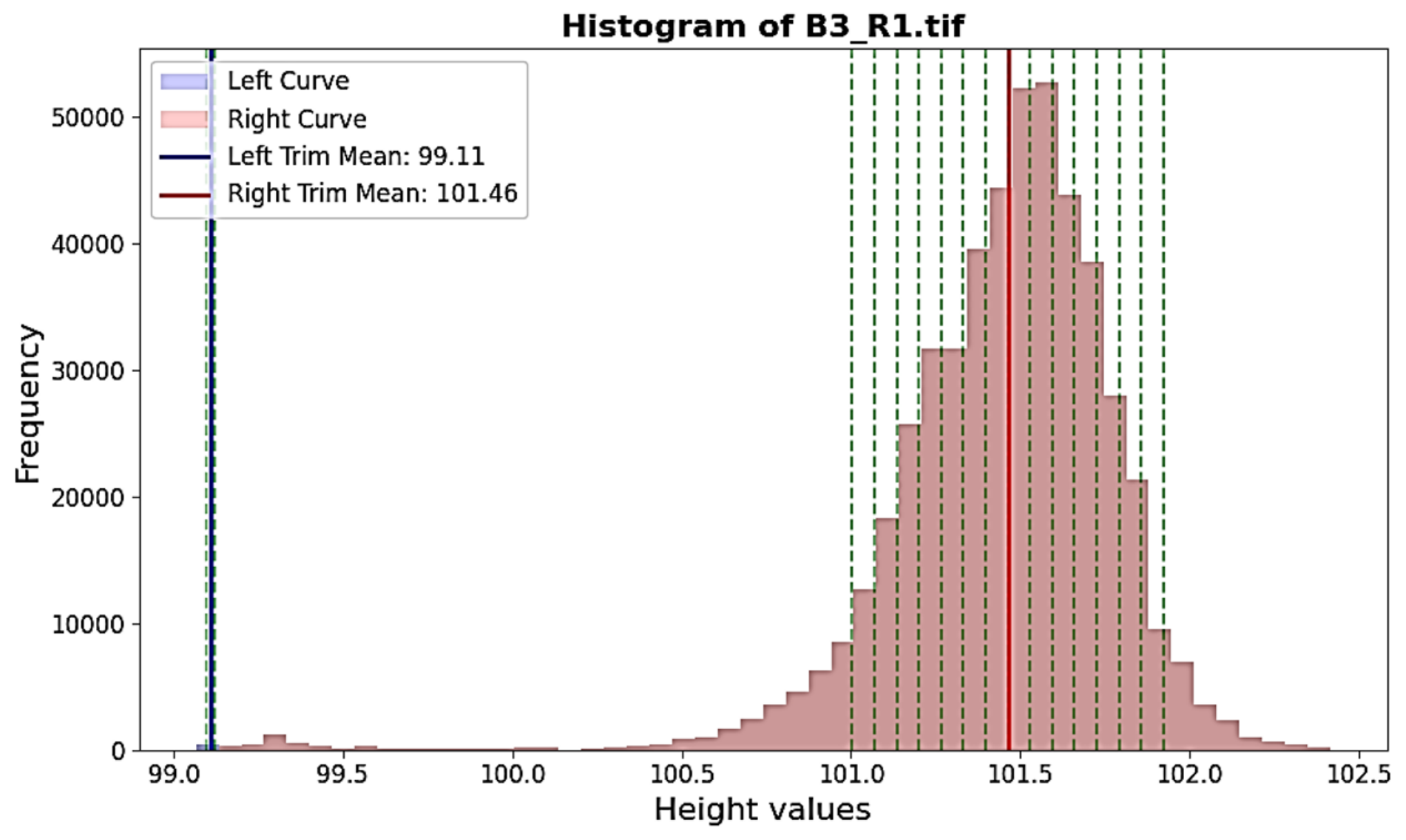}
    \caption{Case 2 - Histogram of Block (B3\_R1) showing a unimodal distribution}
    \label{fig:unimodal}
\end{figure}

\subsection{Case: 2}
In instances where a bimodal distribution was not observed, the histograms exhibited a single peak or unimodal distribution as shown in figure \ref{fig:unimodal}. This scenario typically occurred in plots with dense canopy cover, making it difficult to detect ground-level pixels and rendering the application of a GMM ineffective. To estimate the plant height in these cases, we adopted an alternative approach. We first computed the trimmed mean of the top 30\% of frequency values from the histogram, representing the canopy height. Then, we identified the lowest pixel value within the plot, which corresponds to the ground level. The difference between the trimmed mean and this minimum pixel value provided an accurate estimate of the sugarcane height in plots with high canopy coverage.

\section{Results and Discussion}
\label{sec:typestyle}
\begin{figure}[h!]
    \centering
    \includegraphics[width=1\textwidth]{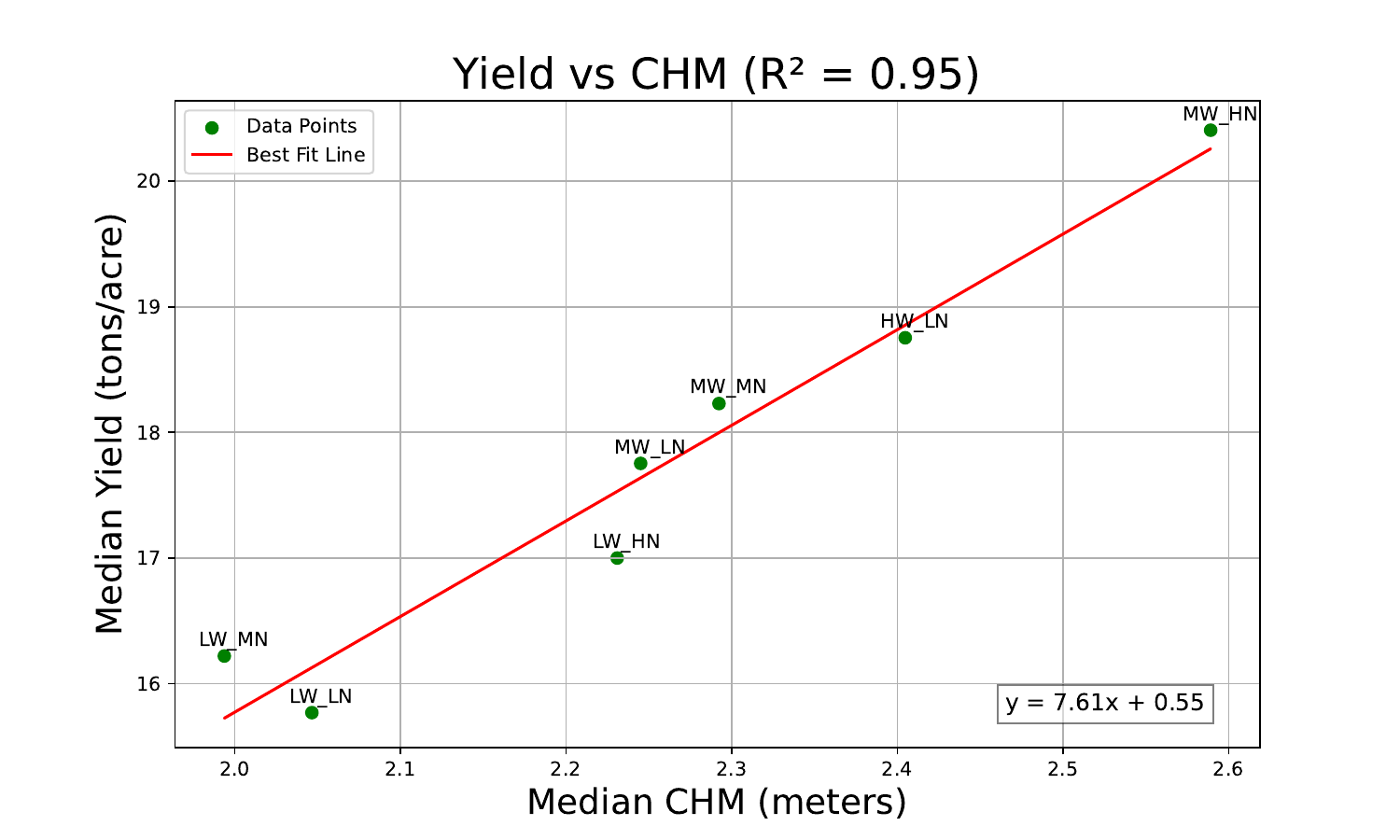}    
    \caption{Regression analysis between sugarcane yield and derived median crop height for each nine class }
    \label{fig:regression}
\end{figure}

In our study, the categorized 62 sugarcane blocks into 9 distinct groups based on the amount of water and nitrogen applied were analyzed. This categorization, detailed in Table \ref{table:water_nitrogen_levels}, facilitated a more granular analysis of how different levels of these inputs affect sugarcane yield. Specifically, we grouped the plots by three water levels (low, medium, high) and three nitrogen levels (low, medium, high), resulting in a comprehensive classification system that allowed for a robust examination of the effects of varying input levels. For each group, we calculated the median cane height and yield, and subsequently conducted a regression analysis to explore the relationship between median crop height and yield across the nine groups. The regression analysis revealed a strong linear relationship between these two variables, as shown in Figure \ref{fig:regression}. The high R² value of 0.95 indicates that 95\% of the variability in yield can be explained by the variability in crop height. The regression model, with the best-fit line equation \( y = 7.61x + 0.56 \), demonstrates a significant direct correlation between crop height and yield. This implies that as crop height increases, the yield also increases proportionally. The model’s high explanatory power underscores the potential of using crop height as a reliable predictor of yield. This finding highlights the importance of optimizing water and nitrogen levels to enhance sugarcane growth and productivity. By understanding and leveraging the relationship between crop height and yield, we can make more informed decisions regarding resource management to improve overall sugarcane production efficiency.

\section{Conclusion}
\label{sec:majhead}
This study effectively demonstrated the strong relationship between sugarcane height and yield at the block level, influenced by varying water and nitrogen treatments. By utilizing UAV-derived Digital Surface Models (DSM) and advanced analytical techniques such as Gaussian Mixture Models (GMM) and regression analysis, we observed a high correlation ($R^{2} = 0.95$) between mean crop height and yield across the different treatment zones. The findings highlight the importance of precise water and nitrogen management in optimizing sugarcane yield, emphasizing that derived sugarcane height can serve as a reliable indicator of yield performance. Using a GMM model to extract surrogate cane height from a single DSM is more effective than the traditional CHM method. Unlike CHM, our approach eliminates the need to capture two separate DSM images—one for the bare ground and another for the canopy cover—to obtain the CHM. Additionally, double DSM capture requires image coregistration, which is not necessary in our method. This research underscores the value of integrating UAV technology with data-driven approaches to enhance the precision and effectiveness of agricultural practices, providing valuable insights for better resource management and decision-making in sugarcane farming.

\section{Future Work}
Future work should focus on expanding the methodologies developed in this study to other regions, validating their generalizability across different agricultural contexts. Additionally, conducting longitudinal studies throughout the growing season would allow for a deeper understanding of how the relationship between crop height and yield evolves over time. Exploring more advanced machine learning models, such as deep learning, could also improve the precision of crop height estimation and yield prediction. Finally, developing user-friendly decision support systems to translate UAV data into actionable insights for farmers would increase the practical utility of this technology, aiding in more informed decision-making for irrigation, fertilization, and harvesting.

\subsubsection{Acknowledgments.} We would like to extend our heartfelt gratitude to CNH Industrial for their invaluable support in supplying the crucial resources and expertise that have significantly contributed to this work. Their dedication to driving innovation and promoting sustainable agricultural practices has been a key factor in the project's advancement. We are equally grateful to Mr. Vikram Bhalla for his generous provision of resources, farm land as well as his hospitality. His unwavering support have played a vital role in the progress of this work, for which we are deeply appreciative.

\newpage
\bibliographystyle{plain}
\bibliography{main}

\end{document}